# Text Relatedness Based on a Word Thesaurus

**George Tsatsaronis**                                                        GBT@IDI.NTNU.NO
*Department of Computer and Information Science*
*Norwegian University of Science and Technology, Norway*

**Iraklis Varlamis**                                                         VARLAMIS@HUA.GR
*Department of Informatics and Telematics*
*Harokopio University, Greece*

**Michalis Vazirgiannis**                                                   MVAZIRG@AUEB.GR
*Department of Informatics*
*Athens University of Economics and Business, Greece*

## Abstract

The computation of relatedness between two fragments of text in an automated manner requires taking into account a wide range of factors pertaining to the meaning the two fragments convey, and the pairwise relations between their words. Without doubt, a measure of relatedness between text segments must take into account both the lexical and the semantic relatedness between words. Such a measure that captures well both aspects of text relatedness may help in many tasks, such as text retrieval, classification and clustering. In this paper we present a new approach for measuring the semantic relatedness between words based on their implicit semantic links. The approach exploits only a word thesaurus in order to devise implicit semantic links between words. Based on this approach, we introduce *Omiotis*, a new measure of semantic relatedness between texts which capitalizes on the word-to-word semantic relatedness measure (*SR*) and extends it to measure the relatedness between texts. We gradually validate our method: we first evaluate the performance of the semantic relatedness measure between individual words, covering word-to-word similarity and relatedness, synonym identification and word analogy; then, we proceed with evaluating the performance of our method in measuring text-to-text semantic relatedness in two tasks, namely sentence-to-sentence similarity and paraphrase recognition. Experimental evaluation shows that the proposed method outperforms every lexicon-based method of semantic relatedness in the selected tasks and the used data sets, and competes well against corpus-based and hybrid approaches.

## 1. Introduction

Relatedness between texts can be perceived in several different ways. Primarily, one can think of lexical relatedness or similarity between texts, which can be easily captured by a vectorial representation of texts (van Rijsbergen, 1979) and a standard similarity measure, like Cosine, Dice (Salton & McGill, 1983), and Jaccard (1901). Such models have had high impact in information retrieval over the past decades. Several improvements have been proposed for such techniques towards inventing more sophisticated weighting schemes for the text words, like for example TF-IDF and its variations (Aizawa, 2003). Other directions explore the need to capture the latent semantic relations between dimensions (words) in the created vector space model, by using techniques of latent semantic analysis (Landauer, Foltz, & Laham, 1998). Another aspect of text relatedness, probably of equal importance, is the semantic relatedness between two text segments. For example, the sentences "*The shares of the company dropped 14 cents*", and "*The business institution's stock slumped 14 cents*" have an obvious semantic relatedness, which traditional measures of text similarity fail to recognize. The motivation of this work is to show that a measure of relatedness between texts, which takes into account both the lexical and the semantic relatedness of word elements, performs better than the traditional lexical matching models, and can handle cases like the one above.

In this paper we propose *Omiotis*[1], a new measure of semantic relatedness between texts, which extends *SR*, a measure of semantic relatedness between words. The word-to-word relatedness measure, in its turn, is based on the construction of semantic links between individual words, according to a word thesaurus, which in our case is WordNet (Fellbaum, 1998). Each pair of words is potentially connected via one or more semantic paths, each one comprising one or more semantic relations (edges) that connect intermediate thesaurus concepts (nodes). For weighting the semantic path we consider three key factors: (a) the semantic path length, (b) the intermediate nodes' specificity denoted by the node depth in the thesaurus' hierarchy, and (c) the types of the semantic edges that compose the path. This triptych allows our measure to perform well in complex linguistic tasks, that require more than simple similarity, such as the SAT Analogy Test[2] that is demonstrated in the experiments. To the best of our knowledge, *Omiotis* is the first measure of semantic relatedness between texts that considers in tandem all three factors for measuring the pairwise word-to-word semantic relatedness scores. *Omiotis* integrates the semantic relatedness in word level with words' statistical information in the text level to provide the final semantic relatedness score between texts.

The contributions of this work can be summarized in the following: 1) a new measure for computing semantic relatedness between words, namely *SR*, which exploits all of the semantic information a thesaurus can offer, including semantic relations crossing parts of speech (POS), while taking into account the relation weights and the depth of the thesaurus' nodes; 2) a new measure for computing semantic relatedness between texts, namely *Omiotis*, that does not require the use of external corpora or learning methods, supervised or unsupervised, 3) thorough experimental evaluation on benchmark data sets for measuring the performance on word-to-word similarity and relatedness tasks, as well as on word analogy; in addition, experimental evaluation on two text related tasks (sentence-to-sentence similarity and paraphrase recognition) for measuring the performance of our text-to-text relatedness measure. Additional contributions of this work are: a) the use of all semantic relations offered by WordNet, which increases the chances of finding a semantic path between any two words, b) the availability of pre-computed semantic relatedness scores between every pair of WordNet senses, which accelerates computation of semantic relatedness between texts and facilitates the incorporation of semantic relatedness in several applications (Tsatsaronis, Varlamis, Nørvåg, & Vazirgiannis, 2009; Tsatsaronis & Panagiotopoulou, 2009).

The rest of the paper is organized as follows: Section 2 discusses preliminary concepts regarding word thesauri, semantic network construction, and semantic relatedness or similarity measures, and summarizes related work on these fields. Section 3 presents the key contributions of our work. Section 4 provides the experimental evaluation and the analysis of the results. Finally, Section 5 presents our conclusions and the next steps of our work.

## 2. Preliminaries and Related Work

Our approach capitalizes on a word thesaurus in order to define a measure of semantic relatedness between words, and expands this measure to compute text relatedness using both semantic and lexical information. In order to facilitate the understanding of our methodology we elaborate on preliminary concepts in this section and present related research approaches.

---

1. *Omiotis* is the Greek word for relatedness or similarity.
2. http://www.aclweb.org/aclwiki/index.php?title=SAT_Analogy_Questions

## 2.1 Word Thesauri and their use in Text Applications

Word thesauri, like WordNet (Fellbaum, 1998) or Roget's International Thesaurus (Morris & Hirst, 1991), constitute the knowledge base for several text-related research tasks. WordNet has been used successfully as a knowledge base in the construction of Generalized Vector Space Models (GVSM) and semantic kernels for document similarity with application to text classification, such as the works of Mavroeidis, Tsatsaronis, Vazirgiannis, Theobald and Weikum (2005), and Basili, Cammisa and Moschitti (2005), and text retrieval, such as the works of Voorhees (1993), Stokoe, Oakes and Tait (2003), and our previous work regarding the definition of a new GVSM that uses word-to-word semantic relatedness (Tsatsaronis & Panagiotopoulou, 2009). Furthermore, the idea of using a thesaurus as a knowledge base in text retrieval has also been proven successful in the case of cross language information retrieval, like for example in the case of the *CLIR* system introduced by Clough and Stevenson (2004). Finally, the exploitation of word thesauri in linguistic tasks, such as Word Sense Disambiguation (WSD) (Ide & Veronis, 1998) has yielded interesting results (Mihalcea & Moldovan, 1999; Tsatsaronis, Vazirgiannis, & Androutsopoulos, 2007; Tsatsaronis, Varlamis, & Vazirgiannis, 2008).

The application of a text relatedness measure to text classification and retrieval tasks should first consider the impact of lexical ambiguity and WSD in the overall performance in these tasks. Sanderson (1994, 2008) concludes that ambiguity in words can take many forms, but new test collections are needed to realize the true importance of resolving ambiguity and embedding semantic relatedness and sense disambiguation in the text retrieval task. In the analysis of Barzilay and El-hadad (1997), and Barzilay, Elhadad and McKeown (2002) the impact of WSD in the performance of text summarization tasks is addressed by considering all possible interpretations of the lexical chains created from text. Similar to this methodology, we tackle word ambiguity by taking into account every possible type of semantic information that the thesaurus can offer, for any given sense of a text word.

From the aforementioned approaches, it is clear that the use of a word thesaurus can offer much potential in the design of models that capture the semantic relatedness between texts, and consequently, it may improve the performance of existing retrieval and classification models under certain circumstances that are discussed in the respective research works (Mavroeidis et al., 2005; Basili et al., 2005; Stokoe et al., 2003; Clough & Stevenson, 2004). The word thesaurus employed in the development of *Omiotis* is WordNet. Its lexical database contains English nouns, verbs, adjectives and adverbs, organized in sets of synonym senses (synsets). Hereafter, the terms *senses*, *synsets* and *concepts* are used interchangeably. Synsets are connected with various links that represent semantic relations between them (i.e., hypernymy / hyponymy, meronymy / holonymy, synonymy / antonymy, entailment / causality, troponymy, domain / domain terms, derivationally related forms, coordinate terms, attributes, stem adjectives, etc.). Several relations cross parts of speech, like the *domain terms* relation, which connects senses pertaining to the same domain (e.g., *light*, as a noun meaning electromagnetic radiation producing a visual sensation, belongs to the domain of *physics*). To the best of our knowledge, the proposed approach is the first that utilizes all of the aforementioned semantic relations that exist in WordNet for the construction of a semantic relatedness measure.

## 2.2 Creating Semantic Networks from Word Thesauri

*Omiotis* is based on the creation of semantic paths between words in a text using the thesaurus' concepts and relations. Early approaches in this field, used gloss words from the respective word definitions in order to build semantic networks from text (Veronis & Ide, 1990). The idea of representing text as a semantic network was initially introduced by Quilian (1969). The expansion of WordNet with semantic relations that cross parts of speech has added more possibilities of semantic network construction from text. More recent approaches to semantic network construction from word thesauri, by Mihalcea, Tarau and Figa (2004) and Navigli (2008), utilize a wide range of WordNet semantic relations instead of the gloss words. These methods outperformed previous methods that used semantic networks in the *all words* WSD tasks of Senseval 2 and 3 for the English language (Palmer, Fellbaum, & Cotton, 2001; Snyder & Palmer, 2004). In this work we adopt the semantic network construction method that we introduced in the past (Tsatsaronis et al., 2007). The method utilizes all of the available semantic relations in WordNet. In the WSD task, the respective method outperformed or matched previous methods that used semantic networks in the *all words* WSD tasks of Senseval 2 and 3 for the English language, and this was largely due to the rich representation that the semantic networks offered. Section 3.1 introduces our semantic relatedness measure.

## 2.3 Measures of Semantic Relatedness

Semantic relatedness between words or concepts has been exploited, in the past, in text summarization (Barzilay et al., 2002), text retrieval (Stokoe et al., 2003; Smeaton, Kelledy, & O'Donnell, 1995; Richardson & Smeaton, 1995) and WSD (Patwardhan, Banerjee, & Pedersen, 2003) tasks. Semantic relatedness measures can be widely classified to dictionary-based[3], corpus-based and hybrid.

Among dictionary-based measures, the measure of Agirre and Rigau (1995) was one of the first measures developed to compute semantic relatedness between two or more concepts (i.e., for a set of concepts). Their measure was based on the density and depth of concepts in the set and on the length of the shortest path that connects them. However, they assume that all edges in the path are equally important.

The measure proposed by Leacock, Miller and Chodorow (1998) for computing the semantic similarity between a pair of concepts takes into account the length of the shortest path connecting them, measured as the number of nodes participating in the path, and the maximum depth of the taxonomy. The measure for two concepts $s_1$ and $s_2$ can be computed as follows:

$$Sim(s_1, s_2) = -log \frac{length}{2 \cdot D} \tag{1}$$

where *length* is the length of the shortest path connecting $s_1$ and $s_2$ and *D* is the maximum depth of the taxonomy used.

Regarding hybrid measures, Resnik's (1995, 1999) measure for pairs of concepts is based on the Information Content (*IC*) of the deepest concept that can subsume both (least common subsumer), and can be considered as a hybrid measure, since it combines both the hierarchy of the used thesaurus, and statistical information for concepts measured in large corpora. More specifically, the semantic similarity for a given pair of concepts $s_1$ and $s_2$, which have $s_0$ as their least common subsumer (i.e., least common ancestor), is defined in the following equation:

$$Sim(s_1, s_2) = IC(s_0) \qquad (2)$$

where the Information Content (IC) of a concept (i.e., $s_0$) is defined as:

$$IC(s_0) = -logP(s_0) \qquad (3)$$

and $P(s_0)$ is the probability of occurrence of the concept $s_0$ in a large corpus.

---

3. Also found in the bibliography as knowledge-based, thesaurus-based, or lexicon-based.

The measure proposed by Jiang and Conrath (1997), is also based on the concept of *IC*. Given two concepts $s_1$ and $s_2$, and their least common subsumer $s_0$, their semantic similarity is defined as follows:

$$Sim(s_1, s_2) = \frac{1}{IC(s_1) + IC(s_2) - 2 \cdot IC(s_0)} \qquad (4)$$

The measure of Lin (1998) is also based on *IC*. Given, again, $s_1$, $s_2$, and $s_0$, as before, the similarity between $s_1$ and $s_2$ is defined as follows:

$$Sim(s_1, s_2) = \frac{2 \cdot IC(s_0)}{IC(s_1) + IC(s_2)} \qquad (5)$$

Hirst and St-Onge (1998) reexamine the idea of constructing lexical chains between words, based on their synsets and the respective semantic edges that connect them in WordNet. The initial idea of lexical chains was first introduced by Morris and Hirst (1991), who defined the lexical cohesion of a passage, based on the cohesion of the lexical chains between the passage's elements, which acted as an indicator for the continuity of the passage's lexical meaning.

We encourage the reader to consult the analysis of Budanitsky and Hirst (2006) for a detailed discussion on most of the aforementioned measures, as well as for more measures proposed prior to the aforementioned. While all these measures use only the noun hierarchy (except from the measure of Hirst and St-Onge), the implementation of several of those measures provided by Patwardhan, Banerjee and Pedersen (2003) in the publicly available *WordNet::Similarity* package can also utilize the verb hierarchy. Still, the relations that cross parts of speech are not considered, as well as other factors discussed in detail in Section 3. In contrast, our measure defines the semantic relatedness between any two concepts, independently of their Part of Speech (POS), utilizing all of the available semantic links offered by WordNet.

More recent works of interest on semantic relatedness, include: the measures of Jarmasz and Szpakowicz (2003), who use Roget's thesaurus to compute semantic similarity, by replicating a number of WordNet-based approaches, the LSA-based measure of Finkelstein et al. (2002), who perform Latent Semantic Analysis (Landauer et al., 1998) to capture text relatedness and can be considered as a corpus-based method, the measure of Patwardhan and Pedersen (2006), who utilize the gloss words found in the words' definitions to create WordNet-based context vectors, the methods of Strube and Ponzetto (2006, 2007a), Gabrilovich and Markovitch (2007), and Milne and Witten (2008) who use Wikipedia to compute semantic relatedness and can also be considered as corpus-based approaches, and the method of Mihalcea, Corley and Strappavara (2006), which is a hybrid method that combines knowledge-based and corpus-based measures of text relatedness. Other recent hybrid measures of semantic similarity are: the measure proposed by Li et al. (2006), who use information from WordNet and corpus statistics collected from the Brown Corpus (Kucera,

Francis, & Caroll, 1967) to compute similarity between very short texts, and the measure for text distance proposed by Tsang (2008), that uses both distributional similarity and ontological/knowledge information to compute the distance between text fragments. Distributional similarity is also used in a supervised combination with WordNet-based approaches (Agirre, Alfonseca, Hall, Kravalova, Pasca, & Soroa, 2009), to produce a supervised measure of semantic relatedness. Li et al. (2006) have created a new data set for their experimental evaluation, which we also use in Section 4 to evaluate our *Omiotis* measure and compare against their approach.

In the following section we formally define *Omiotis* and provide its details, from the creation of the semantic links to the computation of relatedness between words and texts. We give evidence on the measure's complexity and justify our design choices. Finally, we discuss potential applications of the measure on text related tasks.

## 3. Measuring Word-to-Word and Text-to-Text Semantic Relatedness

This section presents the details of *Omiotis*, our measure of text semantic relatedness. The measure capitalizes on the idea of semantic relatedness between WordNet senses, extends it to compute relatedness between words and finally between texts. Since the definition of semantic relatedness ranges from pairs of keyword senses to pairs of texts, *Omiotis* is defined in a way that captures relatedness in every granularity. As a result, it can be applied in a wide range of linguistic and text related tasks such as WSD, word similarity and word analogy, text similarity, and keyword ranking. The key points of the proposed measure are: (a) it constructs semantic links between all word senses in WordNet and pre-computes a relatedness score between every pair of WordNet senses, (b) it computes the semantic relatedness for a pair of words by taking into account the relatedness of their corresponding WordNet senses, and (c) it computes a semantic relatedness score for any two given text segments by extending word-to-word relatedness. Depending on the task, the computation of semantic relatedness can be modified to take into account all or some of the senses of each word, all or some of the words in each text, or to apply additional weights depending on the word importance or sense importance in context. This allows *Omiotis* to be adapted in various text related tasks, without modifying the main process of computing relatedness. In Section 3.1 that follows, we formally define our semantic relatedness measure and in Section 3.2 we provide a detailed justification of our design decisions.

### 3.1 Construct Semantic Links between Words

The first step in measuring the semantic relatedness between two text fragments, is to find the implicit semantic links between the words of the two fragments. Thus, we present a definition of semantic relatedness for a pair of thesaurus concepts, which takes into account the semantic path connecting the concepts, and expands it to measure the relatedness between words. In order to solve the problem of constructing semantic paths between words, we base our approach on our previous method on how to construct semantic networks between words (Tsatsaronis et al., 2007).

### 3.1.1 SEMANTIC NETWORK CONSTRUCTION FROM WORD THESAURI

Figure 1 gives an example of the construction of a semantic network for two words $t_i$ and $t_j$. For simplicity reasons, we assume the construction of a semantic path between senses $S.i.2$ and $S.j.1$ only (Initial Phase), though we could do the same for every possible combination of the two words' senses. Initially, the two sense nodes are expanded using all the semantic links offered by WordNet. The semantic links of the senses, as found in the thesaurus, become the edges and the pointed senses the nodes of the network (Network Expansion). The expansion process is repeated recursively until the shortest [4] path between $S.i.2$ and $S.j.1$ is found. When no path is found from $S.i.2$ to $S.j.1$ then the senses and consequently the words are not semantically related.

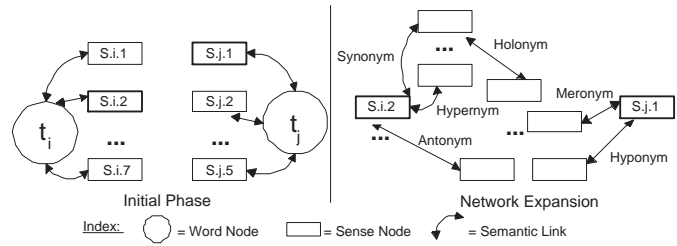

Figure 1: Constructing semantic networks from word thesauri.

### 3.1.2 Semantic Relatedness between a Pair of Concepts

The semantic relatedness for a pair of concepts is measured over the constructed semantic network. It considers the path length, captured by *compactness*, and the path depth, captured by *semantic path elaboration*, which are defined in the following. A measure for WSD based on the idea of *compactness* was initially proposed by Mavroeidis et al. (2005). The original measure used only nouns and the hypernym relation, and is extended in the current work to support all of WordNet's relations and the noun, verb and adjective parts of speech. Here we define a new *compactness* measure (Definition 1) as the core of the *Omiotis* measure.

**Definition 1** *Given a word thesaurus $O$, a weighting scheme for the edges that assigns a weight $w \in (0, 1)$ for each edge, a pair of senses $S = (s_1, s_2)$, and a path $P$ of length $l$ connecting the two senses, the semantic compactness of $S$ ($SCM(S, O, P)$) is defined as: $SCM(S, O, P) = \prod_{i=1}^{l} w_i$, where $w_1, w_2, ..., w_l$ are the path's edges' weights. If $s_1 = s_2$ then $SCM(S, O, P) = 1$. If there is no path between $s_1$ and $s_2$ then $SCM(S, O, P) = 0$.*

Note that *compactness* takes the path length into account and is bound in $[0, 1]$. Higher *compactness* between senses implies higher semantic relatedness. The intuition behind edge types' weighting is that certain types provide stronger semantic connections than others. Considering that the lexicographers of WordNet tend to use some relation types more often than others (we assume that the most used relation types are stronger than the types less used), a straightforward solution is to define edge types' weights in proportion to their frequency of occurrence in WordNet 2.0. The weights assigned to each type using this solution are shown in Table 1 and are in accordance to those found by Song et al. (2004). The table shows the probability of occurrence in WordNet 2.0 for every possible edge type in the thesaurus, in descending order of probability values. A detailed analysis of the choices we made in Definition 1 and in the definitions that follow is performed in Section 3.2.

The depth of nodes that belong to the path also affects term relatedness. A standard means of measuring depth in a word thesaurus is the hypernym/hyponym hierarchical relation for the noun and adjective POS and hypernym/troponym for the verb POS. For the adverb POS the related *stem*

---

4. The details are presented in Algorithm 1.

| WordNet 2.0 Edge Type | Probability of Occurrence |
|---|---|
| hypernym/hyponym | 0.61 |
| nominalization | 0.147 |
| category domain | 0.094 |
| part meronym/holonym | 0.0367 |
| region domain | 0.0238 |
| similar | 0.02 |
| usage domain | 0.016 |
| member meronym/holonym | 0.014 |
| antonym | 0.0105 |
| verb group | 0.01 |
| also see | 0.0091 |
| attribute | 0.00414 |
| entailment | 0.00195 |
| cause | 0.00158 |
| substance meronym/holonym | 0.00089 |
| derived | 0.0003 |
| participle of | $3.4E - 06$ |

Table 1: Probability of occurrence for every edge type in WordNet 2.0.

*adjective* sense can be used to measure its depth. A path with shallow sense nodes is more general compared to a path with deep nodes. This parameter of semantic relatedness between terms is captured by the measure of *semantic path elaboration* introduced in the following definition.

**Definition 2** *Given a word thesaurus $O$ , a pair of senses $S = (s_1, s_2)$, where $s_1, s_2 \in O$ and $s1 \neq s2$, and a path $P = <p_1, p_2, ..., p_l>$ of length $l$, where either $s_1 = p_1$ and $s_2 = p_l$ or $s_1 = p_l$ and $s_2 = p_1$, the semantic path elaboration of the path $(SPE(S, O, P))$ is defined as: $SPE(S, O, P) = \prod_{i=1}^{l} \frac{2d_i d_{i+1}}{d_i + d_{i+1}} \cdot \frac{1}{d_{max}}$, where $d_i$ is the depth of sense $p_i$ according to $O$, and $d_{max}$ the maximum depth of $O$. If $s_1 = s_2$, then $d_1 = d_2 = d$ and $SPE(S, O, P) = \frac{d}{d_{max}}$. If there is no path from $s_1$ to $s_2$ then $SPE(S, O, P) = 0$.*

It is obvious in Definition 2 that a path of length $l$ comprises $l+1$ nodes, thus when $i = l$, $d_{i+1}$ is the last node in the path. Essentially, SPE is the harmonic mean of the two depths normalized to the maximum thesaurus depth. The harmonic mean is preferred over the average of depths, since it offers a lower upper bound and gives a more realistic estimation of the path's depth. *Compactness* and *Semantic Path Elaboration* measures capture the two most important parameters of measuring semantic relatedness between terms (Budanitsky & Hirst, 2006), namely *path length* and *senses depth* in the used thesaurus. We combine these two measures following the definition of the *Semantic Relatedness* between two terms:

**Definition 3** *Given a word thesaurus $O$, and a pair of senses $S = (s_1, s_2)$ the semantic relatedness of $S$ $(SR(S, O))$ is defined as $max_P\{SCM(S, O, P) \cdot SPE(S, O, P)\}$.*

---

**Algorithm 1** Maximum-Semantic-Relatedness($G, u, v, w$)

---

1: **INPUT:** A directed weighted graph $G$, two nodes $u, v$ and a weighting scheme $w : E \rightarrow (0..1)$.

2: **OUTPUT:** The path from $u$ to $v$ with the maximum product of the edges weights.

    *Initialize-Single-Source($G, u$)*

3: **for all** vertices $v \in V_G$ **do**

4:    $d[v] = -\infty$

5:    $\pi[v] = NULL$

6: **end for**

7: $d[u] = 1$

   *Relax($u, v, w$)*

8: **if** $d[v] < d[u] \cdot w(u, v)$ **then**

9:    $d[v] = d[u] \cdot w(u, v)$

10:    $\pi[v] = u$

11: **end if**

   *Maximum-Relatedness($G, u, v, w$)*

12: Initialize-Single-Source($G, u$)

13: $S = \emptyset$

14: $Q = V_G$

15: **while** $v \in Q$ **do**

16:    $s$ = Extract from $Q$ the vertex with the maximum $d$

17:    $S = S \cup s$

18:    **for all** vertices $k \in$ Adjacency List of $s$ **do**

19:       Relax($s, k, w$)

20:    **end for**

21: **end while**

22: return the path following all the ancestors $\pi$ of $v$ back to $u$

---

Given a word thesaurus, there can be more than one semantic path connecting two senses. The senses' *compactness* can take different values for all the different paths. In these cases, we use the path that maximizes the semantic relatedness. For its computation we introduce Algorithm 1, which is a modification of Dijkstra's algorithm (Cormen, Leiserson, & Rivest, 1990) for finding the shortest path between two nodes in a weighted directed graph. In the algorithm, $G$ is the representation of the directed weighted graph given as input (e.g., using adjacency lists), and $V_G$ is the set of all the vertices of $G$. Also, two more sets are used; $S$, which contains all the vertices for which the maximum semantic relatedness has been computed from the source vertex (i.e., from $u$), and $Q$, which contains all the vertices for which the algorithm has not computed yet the maximum relatedness from the source vertex. Furthermore, three tables are used; $d$, which, for any vertex $v$ stores the maximum semantic relatedness found at any given time of the algorithm execution from the source vertex, i.e., $u$ in $d[v]$; $\pi$, which for any vertex $v$ stores its predecessor in $\pi[v]$; and $w$, which stores the edge weights of the graph (e.g., $w[k, m]$ stores the edge weight of the edge that starts from $k$ and goes to $m$).

The algorithm comprises three functions: (a) *Initialize-Single-Source(G, u)*, which initializes tables $d$ and $\pi$, for every vertex $v$ of the graph. More precisely, it sets $d[v] = -\infty$, since the semantic relatedness from the source is unknown at the beginning, and because the algorithm seeks for the maximum semantic relatedness this is initially set to the minimum value (i.e., $-\infty$). It also sets $\pi[v] = NULL$, since at the beginning of the algorithm execution we are not aware yet of the predecessor of any vertex $v$ following the path from the source vertex $u$ to $v$ that results to the maximum semantic relatedness; (b) *Relax(u, v, w)*, which given two vertices, $u$ and $v$ that are directly connected with an edge of weight $w[u, v]$, it updates the value $d[v]$, in case that if we follow the edge $(u, v)$ this results to a higher semantic relatedness for vertex $v$ from the source, compared to the value we have computed up to that time of the algorithm execution; and (c) *Maximum-Relatedness(G, u, v, w)*, which uses the aforementioned functions and executes the Dijkstra's algorithm. The proof of the algorithm's correctness follows in the next theorem.

**Theorem 1** *Given a word thesaurus $O$, an edges weighting function $w : E \rightarrow (0, 1)$, where a higher value declares a stronger edge, and a pair of senses $S(s_s, s_f)$ declaring source $(s_s)$ and destination $(s_f)$ vertices, then the $SCM(S, O, P) \cdot SPE(S, O, P)$ is maximized for the path returned by Algorithm 1, by using the weighting scheme $w'_{ij} = w_{ij} \cdot \frac{2 \cdot d_i \cdot d_j}{d_{max} \cdot (d_i + d_j)}$, where $w'_{ij}$ is the new weight of the edge connecting senses $s_i$ and $s_j$.*

**Proof 1** *We will show that for each vertex $s_f \in V_G$, $d[s_f]$ is the maximum product of edges' weight through the selected path, starting from $s_s$, at the time when $s_f$ is inserted into $S$. From now on, the notation $\delta(s_s, s_f)$ will represent this product. Path $p$ connects a vertex in $S$, namely $s_s$, to a vertex in $V_G - S$, namely $s_f$. Consider the first vertex $s_y$ along $p$ such that $s_y \in V_G - S$ and let $s_x$ be $y$'s predecessor. Now, path $p$ can be decomposed as $s_s \rightarrow s_x \rightarrow s_y \rightarrow s_f$. We claim that $d[s_y] = \delta(s_s, s_y)$ when $s_f$ is inserted into $S$. Observe that $s_x \in S$. Then, because $s_f$ is chosen as the first vertex for which $d[s_f] \neq \delta(s_s, s_f)$ when it is inserted into $S$, we had $d[s_x] = \delta(s_s, s_x)$ when $s_x$ was inserted into $S$.*

*Because $s_y$ occurs before $s_f$ on the path from $s_s$ to $s_f$ and all edge weights are nonnegative and in $(0, 1)$ we have $\delta(s_s, s_y) \geq \delta(s_s, s_f)$, and thus $d[s_y] = \delta(s_s, s_y) \geq \delta(s_s, s_f) \geq d[s_f]$. But both $s_y$ and $s_f$ were in $V - S$ when $s_f$ was chosen, so we have $d[s_f] \geq d[s_y]$. Thus, $d[s_y] = \delta(s_s, s_y) = \delta(s_s, s_f) = d[s_f]$. Consequently, $d[s_f] = \delta(s_s, s_f)$ which contradicts our choice of $s_f$. We conclude that at the time each vertex $s_f$ is inserted into $S$, $d[s_f] = \delta(s_s, s_f)$.*

*Next, to prove that the returned maximum product is the $SCM(S, O, P) \cdot SPE(S, O, P)$, let the path between $s_s$ and $s_f$ with the maximum edge weight product have $k$ edges. Then, Algorithm 1 returns the maximum $\prod_{i=1}^{k} w'_{i(i+1)} = w_{s2} \cdot \frac{2 \cdot d_s \cdot d_2}{d_{max} \cdot (d_s + d_2)} \cdot w_{23} \cdot \frac{2 \cdot d_2 \cdot d_3}{d_{max} \cdot (d_2 + d_3)} \cdot \ldots \cdot w_{kf} \cdot \frac{2 \cdot d_k \cdot d_f}{d_{max} \cdot (d_k + d_f)} = \prod_{i=1}^{k} w_{i(i+1)} \cdot \prod_{i=1}^{k} \frac{2d_i d_{i+1}}{d_i + d_{i+1}} \cdot \frac{1}{d_{max}} = SCM(S, O, P) \cdot SPE(S, O, P).*

### 3.1.3 SEMANTIC RELATEDNESS FOR A PAIR OF TERMS

Based on Definition 3, which measures the semantic relatedness between a pair of senses $S$, we can define the semantic relatedness between a pair of terms $T(t_1, t_2)$ as follows.

**Definition 4** *Let a word thesaurus $O$, let $T = (t_1, t_2)$ be a pair of terms for which there are entries in $O$, let $X_1$ be the set of senses of $t_1$ and $X_2$ be the set of senses of $t_2$ in $O$. Let $S_1, S_2, ..., S_{|X_1| \cdot |X_2|}$ be the set of pairs of senses, $S_k = (s_i, s_j)$, with $s_i \in X_1$ and $s_j \in X_2$. Now the semantic relatedness of $T$ $(SR(T, S, O))$ is defined as:*

$max_{S_k}\{max_P\{SCM(S_k, O, P) \cdot SPE(S_k, O, P)\}\} = max_{S_k}\{SR(S_k, O)\}$
*for all $k = 1..|X_1| \cdot |X_2|$. Semantic relatedness between two terms $t_1, t_2$ where $t_1 \equiv t_2 \equiv t$ and $t \notin O$ is defined as 1. Semantic relatedness between $t_1, t_2$ when $t_1 \in O$ and $t_2 \notin O$, or vice versa, is considered 0.*

For the remaining of the paper, the $SR(T, S, O)$ for a pair of terms will be denoted as $SR(T)$, to ease readability.

### 3.2 Analysis of the SR Measure

In this section we present the rationale behind the Definitions 1, 2, and 3, by providing theoretical and/or experimental evidence for the decisions made on the design of the measure. We illustrate the advantages and disadvantages of the different alternatives using simple examples and argue for our decisions. Finally, we discuss on the advantages of *SR* against previous measures of semantic relatedness.

The list of decisions made for the design of our semantic relatedness measure comprises: a) use of senses in all POS, instead of noun senses only, b) use of all semantic edge types found in WordNet, instead of the IS-A relation only, c) use of edge weights, and d) use of senses' depth as a scaling factor. It is important to mention that measures of semantic relatedness differ from the measures of semantic similarity, which traditionally use hierarchical relations only and ignore all other type of semantic relations. In addition, both concepts differentiate from semantic distance, in the sense that the latter is a metric.

#### 3.2.1 Use all POS Information

Firstly, we shall argue on the fact that the use of all POS in designing a semantic relatedness measure is important, and can increase the coverage of such a measure. The rationale supporting this decision is fairly simple. Current data sets for evaluating semantic relatedness or even semantic similarity measures are restricted to nouns, like for example the Rubenstein and Goodenough 65 word pairs (1965), the Miller and Charles 30 word pairs (1991), and the Word-Similarity-353 collection (Finkelstein et al., 2002). Thus, the experimental evaluation in those data sets cannot pinpoint the caveat of omitting the remaining parts of speech. However, text similarity tasks and their benchmark data sets comprise more than nouns. Throughout the following analysis, the reader must consider that the resulting measure of semantic relatedness among words is destined to be embedded in a text-to-text semantic relatedness, as shown in the next section.

The following two sentences are a paraphrase example taken from the Microsoft Paraphrase Corpus (Dolan, Quirk, & Brockett, 2004) and show the importance of using other POS as well, such as verbs:

> "The **charges** of **espionage** and **aiding** the **enemy** can **carry** the **death penalty**."

> "If **convicted** of the **spying charges** he could **face** the **death penalty**."

Words that appear in WordNet are written in bold and stopwords have been omitted for simplicity[5]. The two sentences have many nouns in common (charges, death, penalty), but there are also pairs of words between these two sentences that can contribute the evidence that these two sentences are

---

a paraphrase. For example **espionage** and **spying** have an obvious semantic relatedness, as well as **enemy** and **spying**. Also, **convicted** and **charges**, as well as **convicted** and **penalty**. This type of evidence would have been disregarded by any measure of semantic relatedness or similarity that uses only the noun POS hierarchy of WordNet. Examples of such measures are: the measure of Sussna (1993), Wu and Palmer (1994), Jiang and Conrath (1997), Resnik (1995, 1999), and the WordNet-based component of the measure proposed by Finkelstein et al. (2002). From this point of view, the decision to use all POS information expands the potential matches found by the measure and allows the use of the measure in more complicated tasks, like paraphrase recognition, text retrieval, and text classification.

### 3.2.2 USE EVERY TYPE OF SEMANTIC RELATIONS

The decision to use all parts of speech in the construction of the semantic graphs, as it was introduced in our previous work (Tsatsaronis et al., 2007), imposes the involvement of all semantic relations instead of merely taxonomic (IS-A) ones. Moreover, this decision was based on evidence from related literature. The work of Smeaton et al. (1995) provides experimental evidence that measuring semantic similarity by incorporating non-hierarchical link types (i.e. part meronym/holonym, member meronym/holonym, substance meronym/holonym) improves much the performance of such a measure. The experimental evaluation was conducted by adopting a small variation of the Resnik's measure (1995).

Hirst and St-Onge (1998) reported that they have discovered several limitations and missing connections in the set of WordNet relations during the construction of lexical chains from sentences for the detection and correction of malapropisms. They provided the following example using the pair of words in bold to report this caveat:

> "School administrators say these same taxpayers expect the schools to provide **child care** and **school** lunches, to integrate immigrants into the community, to offer special classes for adult students,."

The intrinsic connection between the nouns **child care** and **school**, which both exist in WordNet, cannot be discovered by considering only hierarchical edge types. This connection is depicted in Figure 2, which shows the path in WordNet. Our rich semantic representation is able to detect such connections and address problems of the aforementioned type.

### 3.2.3 USE WEIGHTS ON EDGES

The work of Resnik (1999) reports that simple edge counting, which implicitly assumes that links in the taxonomy represent uniform distances, is problematic and is not the best semantic distance measure for WordNet. In a similar direction lie the findings of Sussna (1993), who has performed thorough experimental evaluation by varying edge weights in order to measure semantic distance between concepts. Sussna's findings, revealed that weights on semantic edges are a non-negligible factor in the application of his measure for WSD, and that the best results were reported when an edge weighting scheme was used, instead of assigning each edge the same weight. For all these reasons, we decided to assign a weight on every edge type, and we chose the simple probability of occurrence for each edge type in WordNet, as our edge weighting scheme (see Table 1). This very important factor is absent in several similarity measures proposed in the past, such as in the measures of Leacock et al. (1998), Jarmasz and Szpakowicz (2003) and Banerjee and Pedersen (2003), which are outperformed in experimental evaluation by our measure.

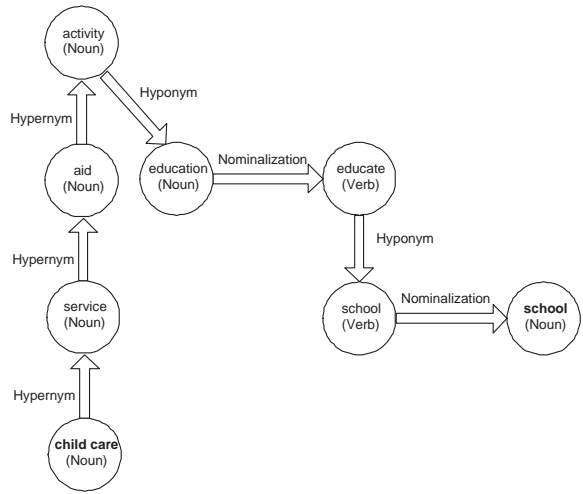

Figure 2: Semantic path from **child care** to **school** following WordNet edges.

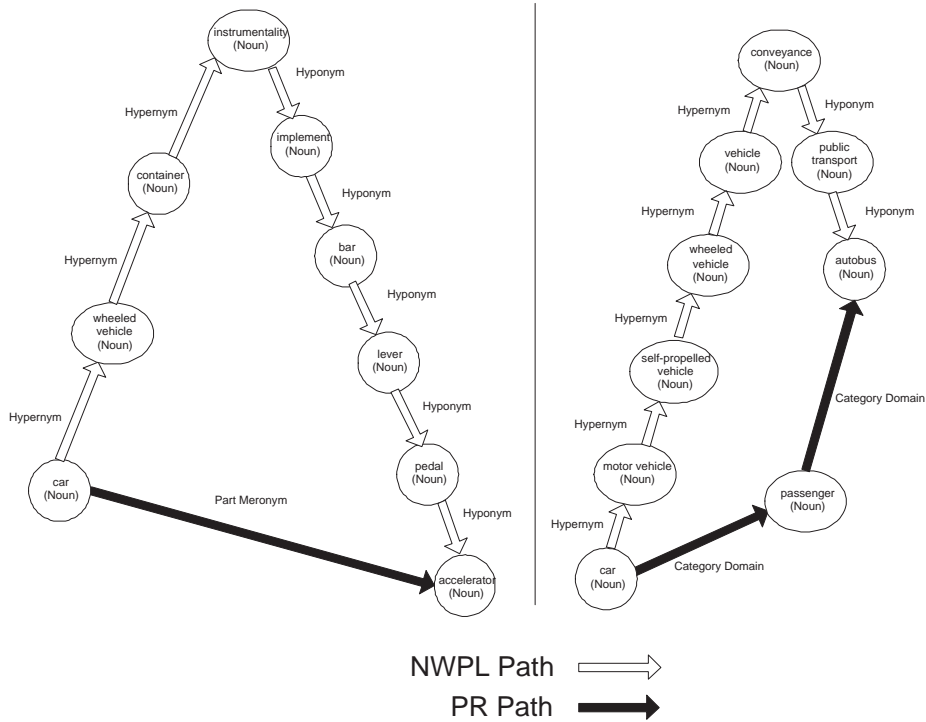

NWPL Path

PR Path

Figure 3: Product Relatedness (PR) and Normalized Weighted Path Length (NWPL) paths for pairs: *car* and *accelerator* (left), *car* and *autobus* (right).

### 3.2.4 USE DEPTH SCALING FACTOR

Our decision to incorporate the depth scaling factor (SPE in Definition 2) in the edge weighting mechanism has been inspired by the thorough experimental evaluation conducted by Sussna (1993), which has provided evidence on the importance of the edge weighting factor in semantic network based measures. According to our experiments on the Miller and Charles data set the Spearman correlation with human judgements was much lower (7 percentage points) when omitting the depth scaling factor than when adopting the SPE factor (see Definition 3).

### 3.2.5 JUSTIFICATION OF *SR* DEFINITIONS

According to Definition 1, the semantic compactness for a pair of concepts is the product of depth-scaled weights of the edges connecting the two concepts. The use of product instead of sum or normalized sum of edges' weights is explained in the following.

Since there might be several paths connecting the two concepts, Definition 3 clearly selects the path that maximizes the product of semantic compactness (SC) and semantic path elaboration (SPE). For simplicity, we ignore the effect of the depth scaling factor (SPE in Definition 2) and consequently, our aim is to find the path that maximizes $\prod_{i=1}^{l} e_i$, where $e_1, e_2, ..., e_l$ are the (non depth-scaled) weights of edges in the path connecting two given concepts. Let us name this less elaborate version of our semantic relatedness measure after *product relatedness* (PR), where $PR(S, O) = max_P \{SCM(S, O, P)\}$. An alternative would have been to define semantic compactness as the normalized sum of the weights in the path, which is: $\frac{\sum_{i=1}^{l} e_i}{l}$. In this case, the semantic relatedness would be measured on the path that maximizes the latter formula, since by nature, semantic relatedness always seeks to find the path that maximizes the connectivity between two concepts. Let us name this alternative after *normalized weighted path length* (NWPL).

In the example of Figure 3, we show how PR and NWPL compute the semantic relatedness for the term pair *car* and *accelerator* (left) and *car* and *autobus* (right). The path that maximizes the respective formulas of PR and NWPL using Algorithm 1 and edge weights in Table 1, is illustrated in Figure 3 using black and white arrows respectively. For the pair *car* and *accelerator* the sum-based formula, normalized against the path length, selects a very large path in this example, with a final computed relatedness of 0.61, which is the weight of the hypernym/hyponym edges. PR finds that the path maximizing the product is the immediate part meronym relation from *car* to *accelerator*, with a computed relatedness of 0.0367, which is the weight of the part meronym edges. The main problem arising with NWPL is the fact that it cannot distinguish among the relatedness between any pair of concepts in the hypernym/hyponym hierarchy of WordNet. In this example, NWPL computes the same relatedness (0.61) between every possible concept pair shown in the top figure. In contrast, PR is able to distinguish most of these pairs in terms of relatedness. More precisely, this behavior of PR is due to the fact that it embeds the notion of the path length, since the computed relatedness decays by a factor in the range $(0, 1)$ for every hop made following any type of semantic relation. Another example, that also shows the importance of considering all WordNet relations, is the one shown on the right part of Figure 3, where NWPL and PR paths have been computed for the term pair *car* and *autobus*. Again, NWPL selects a very large path, and does not incline from the hypernym/hyponym tree.

Clearly, NWPL would rather traverse through a huge path of hypernym/hyponym edges, than following any other less important edge type, which would decrease its average path importance. This behavior creates serious drawbacks: (a) lack of ability to distinguish relatedness among any pair of concepts in the same hierarchy, and (b) large increase of the actual computational cost of Algorithm 1, due to the fact that it will tend not to incline from the hypernym/hyponym hierarchy, even if there is a direct semantic edge (other than hypernym/hyponym) connecting the two concepts, like shown in Figure 3. Furthermore, by conducting experiments with NWPL in the 30 word pairs of Miller and Charles, we discovered that in almost $40\%$ of the cases, NWPL produces the same value of semantic relatedness, equal to $0.61$, being unable to distinguish them and creating many ties. Thus, PR is a better option to use in our measure, as the semantic compactness factor.

Last, but not least, regarding the overall design of *SR*, we should mention that the proposed measure is solely based on the use of WordNet, in contrast to measures of semantic relatedness that use large corpora, such as Wikipedia. Although, such measures, like the ones proposed by Gabrilovich and Markovitch (2007), and Ponzetto and Strube (2007a), provide a larger coverage regarding concepts that do not reside in WordNet, they require the processing of a very large corpora (Wikipedia), which also changes very fast and very frequently. Experimental evaluation in Section 4 shows that our measure competes well against the aforementioned word-to-word relatedness measures in the used data sets. In the following section, we introduce *Omiotis*, the extension of *SR* for measuring text-to-text relatedness.

### 3.3 Omiotis

To quantify the degree to which two text segments semantically relate to each other, we build upon the *SR* measure, which we significantly extend in order to account not only for the terms' semantic relatedness but also for their lexical similarity. This is because texts may contain overly-specialized terms (e.g., an algorithm's name) that are not represented in WordNet. Therefore, relying entirely on the term semantics for identifying the degree to which texts relate to each other would hamper the performance of our approach. On the other hand, semantics serve as complement to our relevance estimations given that different text terms might refer to (nearly-) identical concepts.

To quantify the lexical similarity between two texts, e.g., text $A$ and $B$, we begin with the estimation of the terms' importance weights as these are determined by the standard TF-IDF weighting scheme (Salton, Buckley, & Yu, 1982).

Thereafter, we estimate the lexical relevance, denoted as $\lambda_{a,b}$ between terms $a \in A$ and $b \in B$ based on the harmonic mean of the respective terms' TF-IDF values, given by:

$$\lambda_{a,b} = \frac{2 \cdot TF\_IDF(a, A) \cdot TF\_IDF(b, B)}{TF\_IDF(a, A) + TF\_IDF(b, B)} \qquad (6)$$

Harmonic mean is preferred instead of average, since it provides a more tight upper bound (Li, 2008). This decision is based on the fact that $TF\_IDF(a, A)$ and $TF\_IDF(b, B)$ are two different quantities measuring the qualitative strength of $a$ and $b$ in the respective texts.

Having computed the lexical relevance between text terms $a$ and $b$, we estimate their semantic relatedness, i.e. $SR(a, b)$ as described previously. Based on the estimated lexical relevance and semantic relatedness between pairs of text terms, our next step is to find for every word $a$ in text $A$ the corresponding word $b$ in text $B$ that maximizes the product of semantic relatedness and lexical similarity values as given by Equation 7.

$$b_* = \arg\max_{b \in B}(\lambda_{a,b} \cdot SR(a, b)) \qquad (7)$$

Where $b_*$ corresponds to that term in text $B$, which entails the maximum lexical similarity and semantic relatedness with term $a$ from text $A$.[6] In a similar manner, we define $a_*$, which corresponds to that term in text $A$, which entails the maximum lexical similarity and semantic relatedness with term $b$ from text $B$.

---

6. The function argmax selects the case from the examined ones, that maximizes the input formula of the function.

$$a_* = \arg\max_{a \in A}(\lambda_{a,b} \cdot SR(a,b)) \tag{8}$$

Consequently, we aggregate the lexical and semantic relevance scores for all terms in text $A$, with reference to their best match in text $B$ denoted as shown in Equation 9.

$$\zeta(A,B) = \frac{1}{|A|} \left( \sum_{a \in A} \lambda_{a,b_*} \cdot SR(a,b_*) \right) \tag{9}$$

We do the same for the opposite direction (i.e. from the words of $B$ to the words of $A$) to cover the cases where the two texts do not have an equal number of terms.

Finally, we derive the degree of relevance between texts $A$ and $B$ by combining the values estimated for their terms that entail the maximum lexical and semantic relevance to one another, given by:

$$Omiotis(A,B) = \frac{[\zeta(A,B) + \zeta(B,A)]}{2} \tag{10}$$

Algorithm 2 summarizes the computation of *Omiotis*. Its computation entails a series of steps, the complexity of which is discussed in Section 3.5.

### 3.4 Applications of Semantic Relatedness

In this section we describe the methodology of incorporating semantic relatedness between pairs of words or pairs of text segments, into several applications.

#### 3.4.1 Word-to-Word Similarity

Rubenstein and Goodenough (1965) obtained synonymy judgements from 51 human subjects on 65 pairs of words, in an effort to investigate the relationship between similarity of context and similarity of meaning (synonymy). Since then, the idea of evaluating computational measures of semantic relatedness by comparing against human judgments on a given set of word pairs, has been widely used, and even more data sets were developed. The proposed measure of semantic relatedness between words (*SR*), introduced in Definition 4, can be used directly in such a task, in order to evaluate the basis of *Omiotis* measure, which is the measurement of word-to-word semantic relatedness. The application is straightforward: Let $n$ be all pairs of words in the used word similarity data set. Then, the semantic relatedness for every pair is computed, using $SR(T,S,O)$ as defined in 4. The computed values are sorted in a descending order, and the produced ranking of similarities is compared against the "gold standard" ranking of humans, using Spearman correlation. The scores can be used to compute Pearson's product moment correlation. Additional measures of semantic relatedness can be compared against each other by examining the respective correlation values with human judgements.

#### 3.4.2 SAT Analogy Tests

The problem of identifying similarities in word analogies among pairs of words is a difficult problem and it has been standardized as a test for assessing the human ability for language understanding,

---

**Algorithm 2** Omiotis(A,B, Sem, Lex )

---

1: **INPUT:** Two texts A and B, comprising m and n terms each ($a$ and $b$ are terms from A and B respectively),

a semantic relatedness measure $Sem : SR(a, b) \rightarrow (0..1)$,

a weighting scheme of term importance in a text $Lex : TF\_IDF(a, A) \rightarrow (0..1)$

2: **OUTPUT:** Find the pair of terms that maximizes the product of Sem and Lex values.

*Compute-Zeta(A,B)*

3: $sum(A) := 0$

4: **for all** terms $a \in A$ **do**

5:    $b_* := NULL$

6:    $TempZeta := 0$

7:    **for all** terms $b \in B$ **do**

8:       $\lambda_{a,b} = \frac{2 \cdot Lex(a,A) \cdot Lex(b,B)}{Lex(a,A) + Lex(b,B)}$

9:       **if** $TempZeta < \lambda_{a,b} \cdot Sem(a, b)$ **then**

10:          $TempZeta = \lambda_{i,j} \cdot Sem(a, b)$

11:          $b_* = b$

12:       **end if**

13:    **end for**

14:    $sum(A) := sum(A) + TempZeta$

15: **end for**

16: $Zeta(A, B) := sum(A)/|A|$

*Compute-Omiotis(A,B)*

17: $Omiotis(A, B) := \frac{Zeta(A,B) + Zeta(B,A)}{2}$

---

under the scope of the well known SAT analogy tests (Scholastic Aptitude Tests). SAT tests are used as admission tests by universities and colleges in the United States. The participants' aim is to locate out of the five given word pairs the one that presents the most similar analogy to the target pair.

Although it is difficult for machines to model the human cognition of word analogy, several approaches exist in the bibliography that attempt to tackle this problem. Previous approaches can be widely categorized into: corpus-based, lexicon-based and hybrid. Some examples of corpus-based are the approaches of Turney (2008b) and Bicici and Yuret (2006). Examples of lexicon-based approaches, are those of Veale (2004) and the application of the lexicon-based measure by Hirst and St-Onge (1998) in SAT, that can be found in the work of Turney (2006). Hybrid approaches are applied in SAT, through the application of the measures of Resnik (1995) and Lin (1998) that can also be found in the work of Turney (2006).

In order for the reader to understand the difficulty of answering SAT questions, we must point out that the average US college applicant scores 57% (Turney & Littman, 2005), while the top corpus-based approach scores 56.1% (Turney, 2006), the top lexicon-based scores 42% (Veale, 2004) and the top hybrid scores 33.2% (Resnik, 1995).

Another way of categorizing the approaches that measure semantic similarity in analogy tasks is to distinguish among attributional and relational similarity measures (Gentner, 1983).[7] Representative approaches of the first category are lexicon-based approaches, while paradigms of relational similarity measures can be found in approaches based on Latent Relational Analysis (LRA) (Turney, 2006). It is of great interest to point out that LRA-based approaches, like the LRME algorithm proposed recently by Turney (2008a), are superior to attributional similarity approaches in discovering word analogies. This fact is also supported by the experimental findings of Turney (2006). Without doubt, relational similarity approaches may perform better in the SAT analogy task, but still, as shown later in the experiments we conducted in other applications, like paraphrase recognition, the lexicon-based measures can outperform LRA-based approaches in such tasks.

Semantic relatedness (*SR*) between words, as applied in *Omiotis*, can be exploited to solve the word analogy task. The aim of word analogy is, given a pair of words $w_1$ and $w_2$, to identify the series of semantic relations that lead from $w_1$ to $w_2$ (semantic path). In the SAT test, the target pair $(w_1, w_2)$ and candidate word pairs $(w_{1k}, w_{2k})$, with $k$ usually being from 1 to 5, are processed in order to find each pair's analogy. The aim is to locate the pair $k$, which exposes maximum similarity to $w_1$ and $w_2$. A straightforward method to choose among the 5 candidate pairs is to employ two criteria: At first, the $k$ analogies to the analogy of the target pair can be compared, and then the candidate that shows by far the most similar analogy can be selected. However, when the most similar analogy is not obvious, all the 6 pairs may be examined together in order for the slightest differences that lead to the correct answer to be discovered. We attempt to model human cognition of this task using *SR* in a two fold manner: (a) we measure *SR* to capture the horizontal analogy between the given pair and the possible candidate pairs, and (b) we measure *SR* to capture the vertical analogy between the given pair and the possible candidate pairs. These two aspects are covered by the following Equations 11 to 13. To capture the horizontal analogy between a pair of words and a candidate pair, we measure the difference of the *SR* score between the two words respectively as shown:

$$s_1(w_{1k}, w_{2k}) = 1 - |SR(w_1, w_2) - SR(w_{1k}, w_{2k})| \qquad (11)$$

Essentially, $s_1$ expresses the horizontal analogy of the candidate pair $(w_{1k}, w_{2k})$ with the given pair $(w_1, w_2)$. Similarly, we capture the notion of the vertical analogy between the two pairs by computing the difference of the *SR* scores among the two pairs' words, as follows:

$$s_2(w_{1k}, w_{2k}) = 1 - |SR(w_1, w_{1k}) - SR(w_2, w_{2k})| \qquad (12)$$

Finally, we rank candidates depending on the combined vertical and horizontal analogy they have with the given pair, according to the following equation:

$$s(w_{1k}, w_{2k}) = \frac{s_1(w_{1k}, w_{2k}) + s_2(w_{1k}, w_{2k})}{2} \qquad (13)$$

Eventually, we select the candidate pair with the maximum combined score, taking into account both aspects (horizontal and vertical) of analogy between the given and the candidate pairs.

The intuition behind the selection of the these two scores for handling the SAT test, is the following. The order of the words in the pairs (both target and candidates) is not random. Usually, given a pair $(w_1, w_2)$, and the candidate pairs $(w_{1k}, w_{2k})$ the test is solved if one can successfully find the analogy: $w_{1k}$ is to $w_{2k}$ what $w_1$ is to $w_2$. From this perspective, $s_1$ and $s_2$ try to find the candidate pair that best aligns with the target pair. Figure 4 illustrates these two types of analogies (horizontal and vertical) for an example SAT question.

---

7. Two objects, X and Y, are attributionally similar when the attributes of X are similar to the attributes of Y. Two pairs, A:B and C:D, are relationally similar when the relations between A and B are similar to the relations between C and D.

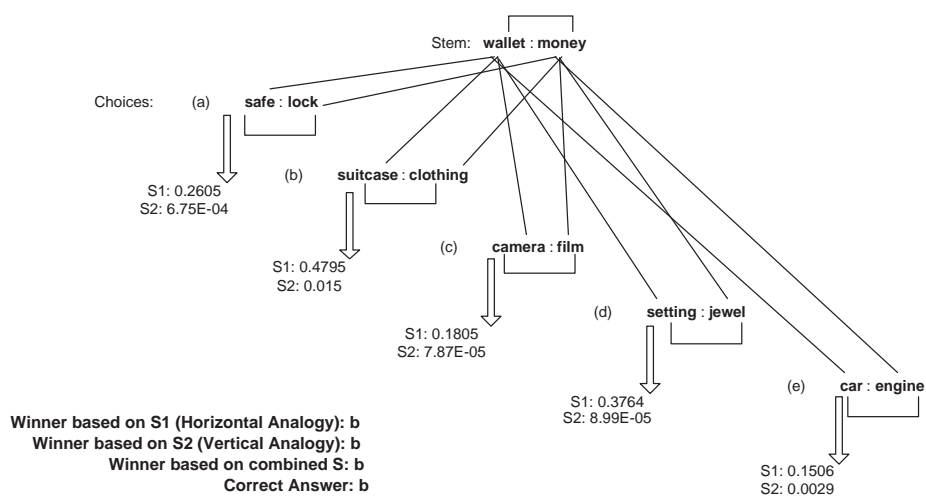

Figure 4: Example of computing the Semantic Relatedness measure (*SR*) in a given Scholastic Aptitude Test (SAT) question.

In order to motivate more our selection of $s_1$ and $s_2$ for answering SAT questions, we will discuss in more detail how these two quantities pertain to the concepts of *strength* and *type* of the relations between a pair of SAT words. Turney (2006) describes a method for comparing the relations between candidate word pairs and the stem word pair, in which he utilizes the *type* of the relation connecting the words in each pair and finally selects the pair that best matches the *type* of the relation connecting the words in the stem pair. Though we do not explicitly examine the label of the edges connecting the words in each pair, implicitly we do so by computing *SR* between them. Since our weighting of the WordNet edges is fine grained, and distinguishes every type of semantic relation in WordNet, instead of labels, we are using edge weights. *SR* definition can provide a fine grained distinguishment between two pairs of words, depending on the *types* of the edges connecting the words respectively, which is expressed by their weights, and also taking into account other factors, like the depth of the nodes comprising their connecting path inside the thesaurus. Besides $s_1$, which attempts to capture the aforementioned properties between word pairs, $s_2$ attempts the same between the words of the same order among two word pairs (i.e., the first word from the first pair, with the second word from the second pair). This forms an attempt to capture how aligned are two word pairs, according to their *SR* values between their words.

### 3.4.3 PARAPHRASE RECOGNITION AND SENTENCE-TO-SENTENCE SIMILARITY

Performance of applications relying on natural language processing may suffer from the fact that the processed documents might contain lexically different, yet semantically related, text segments. The task of recognizing synonym text segments, which is better known as paraphrase recognition, or detection, is challenging and difficult to solve, as shown in the work of Pasca (2005). The task itself is important for many text related applications, like summarization (Hirao, Fukusima, Oku-mura, Nobata, & Nanba, 2005), information extraction (Shinyama & Sekine, 2003) and question answering (Pasca, 2003). We experimentally evaluate the application of *Omiotis* in the paraphrasing detection task (Section 4.2), using the Microsoft Research Paraphrase Corpus (Dolan et al., 2004). The application of *Omiotis* in paraphrase detection is straightforward: given a pair of text segments, we compute the *Omiotis* score between them, using Equation 10 and Algorithm 2. Higher values of *Omiotis* for a given pair denote stronger semantic relation between the examined text segments. The task is now reduced to define a threshold, above which an *Omiotis* value can be considered as a determining sign of a paraphrasing pair. In the experimental evaluation of *Omiotis*, we explain in detail how we have selected this threshold for the paraphrase recognition task.

In a similar manner, by using Equation 10 and Algorithm 2, the semantic relatedness scores for pairs of sentences can be computed. For this task, we are using the data set of Li et al. (2006) to evaluate *Omiotis*, comprising 30 sentence pairs, for which human scores are provided. In Section 4 we describe in detail the experimental set up.

### 3.5 Complexity and Implementation Issues

The computation of *Omiotis* entails a series of steps, the complexity of which is strongly related to its base measure of Semantic Relatedness (*SR*). Primarily, given two words, $w_1$ and $w_2$ the construction time of the semantic network used to compute *SR* according to Algorithm 1, has been proven to be $O(2 \cdot k^{l+1})$ (Tsatsaronis et al., 2007), where $k$ is the maximum branching factor of the used thesaurus nodes and $l$ is the maximum semantic path length in the thesaurus. Once the semantic network is constructed, the complexity of Algorithm 1 is reduced to the standard time complexity cost of Dijkstra's algorithm. Using Fibonacci heaps, it is possible to alleviate the computational burden of Dijkstra and further improve time complexity. In the semantic network, Dijkstra takes $O(nL + mD + nE)$, where $n$ is the number of nodes in the network, m the number of edges, $L$ is the time for insert, $D$ the time for decrease-key and $E$ the time for extract-min. If Fibonacci heaps are used then $L = D = O(1)$ and the cost of extract-min is $O(logn)$, thus significantly reducing the cost of execution. This whole procedure is repeated $2 \times n_1 \times n_2$ times for the computation of *Omiotis* between two documents $d_1$ and $d_2$ having in total $n_1$ and $n_2$ distinct words respectively.

From the aforementioned, it is obvious that the computation of *Omiotis* is not cheap in general. For this purpose, and in order to improve the system's scalability, we have pre-computed and stored all *SR* values between every possible pair of synsets in a RDBMS. This is a one-time computation cost, which dramatically decreases the computational complexity of *Omiotis*. The database schema has three entities, namely *Node*, *Edge* and *Paths*. *Node* contains all WordNet synsets. *Edge* indexes all edges of the WordNet graph adding weight information for each edge computed using the *SR* measure. Finally, *Paths* contains all pairs of WordNet synsets that are directly or indirectly connected in the WordNet graph and the computed relatedness. These pairs were found by running a Breadth First Search (BFS) starting from all WordNet roots for all POS. Table 2 provides statistical information for the RDBMS which exceeds 220 Gbytes in size. Column 1 indicates the number of distinct synsets examined, column 2 shows the total number of the edges, and column 3 depicts the number of the connected synsets (by at least one path following the offered WordNet edges). The current implementation takes advantage of the database structures (indices, stored procedures etc) in order to decrease the computational complexity of *Omiotis*. The following example is indicative of the complexity of *SR* computation. The average number of senses per term is between 5 and 7 (depending on the POS). For a pair of terms of known POS, we perform $\frac{n^2}{2}$ ($n \simeq 6$) combinations and for each pair of synsets we compute the similarity as presented in Definition 3. When these similarities are pre-computed, the time required for processing 100 pairs of terms is $\simeq 1$ sec, which makes the computation of *Omiotis* feasible and scalable. As a proof of concept, we have developed an on-line version of the *SR* and the *Omiotis* measures[8], where the user can test the term-to-term and sentence-to-sentence semantic relatedness measures (Tsatsaronis et al., 2009).

| Distinct Synsets | Total Edges | Connected Synset Pairs |
|:---:|:---:|:---:|
| 115,424 | 324,268 | 5,591,162,361 |

Table 2: Statistics of the WordNet graph in the implemented database.

## 4. Experimental Evaluation

The experimental evaluation of *Omiotis* is two-fold. First, we test the performance of the word-to-word semantic relatedness measure (*SR*), in which *Omiotis* is based, in three types of tasks: (a) word-to-word similarity and relatedness, (b) synonym identification, and (c) Scholastic Aptitude Test (SAT). Second, we evaluate the performance of *Omiotis* in two tasks: (a) sentence-to-sentence similarity, and (b) the paraphrase recognition task.

### 4.1 Evaluation of the Semantic Relatedness (SR) Measure

For the evaluation of the proposed semantic relatedness measure between two terms we experimented on three different categories of tests. The first category comprises data sets that contain word pairs, for which human subjects have provided similarity scores or relatedness scores. The provided scores create a ranking of the word pairs, from the most similar to the most irrelevant. We evaluate the performance of measures, by computing the correlation between the list of the human rankings and the list produced by the measures. In this task, we evaluate the performance of *SR* in three benchmark data sets, namely the Rubenstein and Goodenough 65 word pairs (1965) (R&G), and the Miller and Charles 30 word pairs (1991) (M&C), for which humans have provided similarity scores, and, also, in the Word-Similarity-353 collection (Finkelstein et al., 2002) (353-C), which comprises 353 word pairs, for which humans have provided relatedness scores.

The second category of experiments comprises synonym identification tests. In these tests, given an initial word, the most appropriate synonym word must be identified among the given options. In this task we evaluate the performance of *SR* in the TOEFL data set, comprising 80 multiple choice synonym questions, and the ESL data set, comprising 50 multiple choice synonym questions questions.[9]

The third category of experiments is based on the Scholastic Aptitude Test (SAT) questions. In SAT, given a pair of words, the most relevant pair among five other given pairs must be selected. This task is based on word analogy identification. The evaluation data set comprises 374 test questions.

---

8. Publicly available at http://omiotis.hua.gr
9. http://www.aclweb.org/aclwiki/index.php?title=TOEFL_Synonym_Questions
   http://www.aclweb.org/aclwiki/index.php?title=ESL_Synonym_Questions_(State_of_the_art)

| Category | Method | R&G | | M&C | |
|---|---|---|---|---|---|
| | | Spearman's $\rho$ | Pearson's $r$ | Spearman's $\rho$ | Pearson's $r$ |
| **Lexicon-based** | **HS** | $0.745^\S$ | $0.786^\S$ | $0.653^\S$ | $0.744^\ddagger$ |
| | **LC** | $0.785^\ddagger$ | $0.838$ | $0.748^\ddagger$ | $0.816$ |
| | **JS** | N/A | $0.818^\dagger$ | N/A | $0.878$ |
| **Corpus-based** | **GM** | $0.816^\dagger$ | N/A | $0.723^\ddagger$ | N/A |
| | **WLM** | $0.64^\S$ | N/A | $0.70^\S$ | N/A |
| | **SP** | N/A | $0.52^\S$ | N/A | $0.47^\S$ |
| | **IS-A SP** | N/A | $0.70^\S$ | N/A | $0.69^\ddagger$ |
| **Hybrid** | **JC** | $0.709^\S$ | $0.781^\S$ | $0.805$ | $0.85$ |
| | **L** | $0.77^\ddagger$ | $0.818^\dagger$ | $0.767^\dagger$ | $0.829$ |
| | **R** | $0.7485^\S$ | $0.778^\S$ | $0.737^\ddagger$ | $0.774^\dagger$ |
| | **HR** | $0.817$ | N/A | $0.904$ | N/A |
| | **SR** | **0.8614** | **0.876** | **0.856** | **0.864** |

Table 3: Spearman's and Pearson's correlations for the Rubenstein and Goodenough (R&G) and Miller and Charles (M&C) data sets. Confidence levels: $^\dagger$=0.90, $^\ddagger$=0.95, $^\S$=0.99

### 4.1.1 EVALUATION ON SEMANTIC SIMILARITY AND RELATEDNESS

For the first category of experiments, we compared our measure against ten known measures of semantic relatedness: Hirst and St-Onge (1998) (HS), Jiang and Conrath (1997) (JC), Leacock et al. (1998) (LC), Lin (1998) (L), Resnik (1995, 1999) (R), Jarmasz and Szpakowicz (2003) (JS), Gabrilovich and Markovitch (2007, 2009) (GM), Milne and Witten (2008) (WLM), Finkelstein et al. (2002) (LSA), Hughes and Ramage (2007) (HR), and Strube and Ponzetto (2006, 2007a) (SP). For the measure of Strube and Ponzetto we have also included the results of a version of the measure that is only based on IS-A relations (Ponzetto & Strube, 2007b) (IS-A SP). For each measure, including our own measure (*SR*), we have computed both the Spearman rank order correlation coefficient ($\rho$) and the Pearson product-moment correlation coefficient ($r$), with $\rho$ being derived from $r$, since for the computation of $\rho$ the relatedness scores are transformed into rankings. Both correlation coefficients are computed based on the relatedness scores and rankings provided by humans in all three data sets (the relatedness scores create a ranking of the pairs of words, based on their similarity). For the measures HS, JC, LC, L and R, the rankings and the relatedness scores of the word pairs for the R&G and the M&C data sets, are given in the work of Budanitsky and Hirst (2006). For the JS measure, the $r$ value is given in the work of Jarmasz and Szpakowicz (2003) for the R&G and the M&C data sets, and the $\rho$ value is given in the work of Gabrilovich and Markovitch (2007). For the GM measure the $\rho$ values are given in the work of Gabrilovich and Markovitch (2007). For the WLM measure the $\rho$ values are given in the work of Milne and Witten (2008). For the LSA method the $\rho$ value is given in the work of Gabrilovich and Markovitch (2007), only for the 353-C data set. For the HR measure the $\rho$ values are given in the work of Hughes and Ramage (2007). Finally, for the SP measure the $r$ values are given in the work of Ponzetto and Strube (2007a), and for the IS-A SP are given in the work of Ponzetto and Strube (2007b).

In Table 3 we show the values of $\rho$ and $r$ for the R&G and the M&C data sets and for *SR* and the compared measures. The human scores for all pairs of words for the two data sets can be found in the analysis of Budanitsky and Hirst (2006). Note that the M&C data set is a subset of the R&G data set. In some cases, the computation of $\rho$ or $r$ was not feasible, due to missing information regarding the detailed rankings or relatedness scores for the respective measures. In these cases the table has the entry $N/A$. Also the LSA measure is omitted in this table because $\rho$ and $r$ were not reported in the literature for these two data sets. We have also conducted a statistical significance test on the difference between *SR* correlations and the respective correlations of the compared measures, using Fisher's z-transformation (Fisher, 1915). For each reported number, the symbol § indicates that the difference between the correlation produced by *SR* and the respective measure is statistically significant at the 0.99 confidence level ($p < 0.01$). The symbol ‡ indicates the same at the 0.95 confidence level ($p < 0.05$) and, finally, the symbol † indicates statistical significance of the correlations' difference at the 0.90 confidence level ($p < 0.10$). In cases when the difference is not statistically significant in any of those confidence levels, there is no indicating symbol.

In Table 4 we show the values of $\rho$ and $r$ for the 353-C data set. The reason we present the results of the experiments in the 353-C data set in another table than the respective results of the R&B and M&C data sets is that this collection focuses on the concept of *semantic relatedness*, rather than on the concept of *semantic similarity* (Gabrilovich & Markovitch, 2007). Relatedness is more general concept than similarity, as argued in the analysis of Budanitsky and Hirst (2006). Thus, it can be argued that the humans in the 353-C thought differently when scoring, compared to the case of the R&B and M&C data sets. The detailed human scores for the 353-C data set are made available with the collection[10]. The measures L, JC and HS are omitted, because no information was available for computing $\rho$ or $r$ values. As a further remark regarding the 353-C collection, we need to add the fact that there are cases where the inter-judge correlations may fall below 65%, while R&B and M&C data sets have inter-judge correlations between 0.88 and 0.95. Again, statistical significance tests have been conducted using the Fisher's z-transformation, regarding the difference of *SR* correlations and the correlations of the compared measures. The used symbols that indicate the level of the statistical significance are the same as previously. With regards to the reported correlations for the R&G and M&C data sets, it is shown that *SR* performs very well, since in the majority of the cases *SR* has higher correlation compared to the other measures of semantic relatedness or similarity of any category (knowledge-based, corpus-based or hybrid). In the R&G data set *SR* reports the highest $\rho$ and $r$ correlations. In the M&C data set *SR* has the second highest $\rho$ correlation. The HR measure has the highest $\rho$ correlation, but in the R&G and 353-C *SR* outperforms HR. The differences between *SR* and HR are not statistically significant in any of the two examined data sets. Also, in the M&C data set *SR* has the second $r$ correlation with the JS reporting the highest, but JS is outperformed by *SR* in the R&G and 353-C data sets. In the case of the M&C data set, the difference between *SR* and JS is not statistically significant, but *SR* outperforms JS in the R&G and the 353-C data sets, with statistically significant difference in the reported correlations. Another important conclusion from the results, is the fact that the IS-A SP measure performs better than the SP measure. This is mainly due to the fact that for the computation of the similarity values in such data sets, the inclusion of only IS-A relations is much more reasonable (Ponzetto & Strube, 2007b). The differences in their results (SP and IS-A SP) motivate even more our *SR* measure, since we take the best of both worlds, i.e., we weigh IS-A relations high, and fall back to other relations if necessary.

---

10. `http://www.cs.technion.ac.il/~gabr/resources/data/wordsim353/`

| Category | Method | 353-C | |
|---|---|---|---|
| | | Spearman's $\rho$ | Pearson's $r$ |
| Lexicon-based | **LC** | N/A | $0.34^\S$ |
| | **JS** | $0.55^\ddagger$ | N/A |
| Corpus-based | **GM** | $0.75^\S$ | N/A |
| | **WLM** | $0.69^\ddagger$ | N/A |
| | **LSA** | $0.56^\dagger$ | N/A |
| | **SP** | N/A | $0.49^\S$ |
| Hybrid | **R** | N/A | $0.34^\S$ |
| | **HR** | $0.552^\ddagger$ | N/A |
| | **SR** | **0.61** | **0.628** |

Table 4: Spearman's and Pearson's correlations for the 353 word pairs (353-C) data set. Confidence levels: $^\dagger$=0.90, $^\ddagger$=0.95, $^\S$=0.99

Regarding the 353-C data set, the results in Table 4 show that *SR* again performs well, with the top performers being the Wikipedia-based approaches (Gabrilovich & Markovitch, 2009; Milne & Witten, 2008). The difference between them is statistically significant, but we should note that *SR* outperforms both GM and WLM in the R&G and M&C data sets, with statistically significant difference as well. Partly, this difference in the performance of *SR* compared to GM and WLM can be explained as follows: the GM measure considers words in context (Gabrilovich & Markovitch, 2009), and thus inherently performs word sense disambiguation; in contrast, *SR* takes as input a pair of words, lacks context, and is based only on the information existing in WordNet, which, especially for several of the cases in the 353-C data set, creates a disadvantage (e.g., in the word pair *Arafat* and *Jackson*, there are 11 different entries for the second word in WordNet). The same holds for the WLM measure. Another reason for this difference in performance is the coverage of WordNet. In several cases, one or both of the two words in the 353-C data set comprising a pair, do not exist in WordNet (e.g., the football player *Maradona*). However, as expected, and also shown in the experimental analysis of *Omiotis* that follows, when context is considered, the proposed semantic relatedness measure performs better (the reader may wish to consult Table 9, where for a subset of the R&G data set that contains the full definitions of the words, the correlations of *Omiotis* with the human judgements are the top found among the compared approaches).

To visualize the performance of our measure in a more comprehensible manner, we also present in Figure 5 the relatedness values given by humans for all pairs in the *R&G* and *M&C* data sets, in increasing order of value (left side) and the respective values for these pairs produced using *SR* (right side). Note that the x-axis in both charts begins from the least related pair of terms, according to humans, and continues up to the most related pair of terms. The y-axis in the left chart is the respective humans' rating for each pair of terms. The right figure shows *SR* for each pair. A closer look on Figure 5 reveals that the values produced by *SR* (right figure) follow a pattern similar to that of the human ratings (left figure).

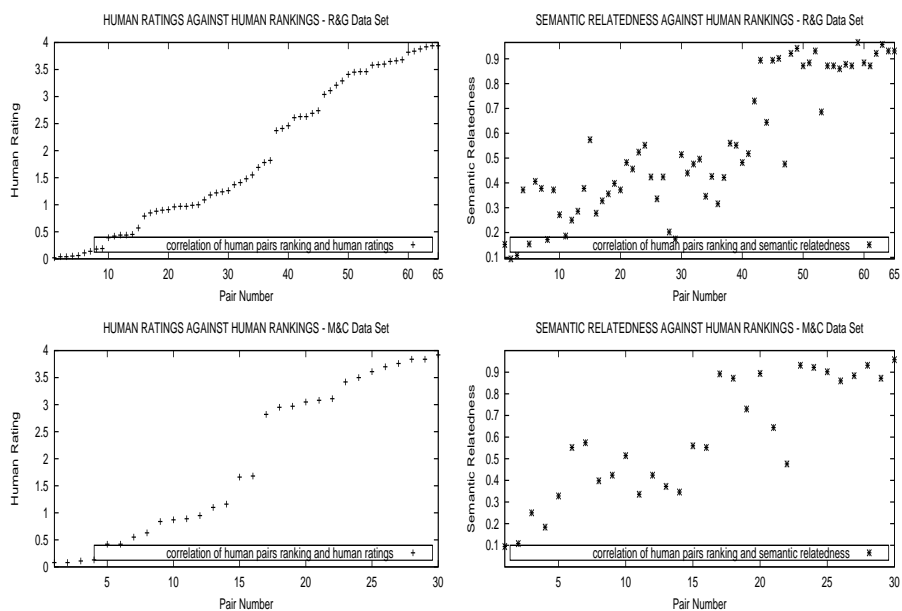

Figure 5: Correlation between human ratings and Semantic Relatedness measure (*SR*) in the Rubenstein and Goodenough (R&G) and Miller and Charles (M&C) data sets.

### 4.1.2 Evaluation on Synonym Identification

For the synonym identification task we are using the TOEFL 80 questions data set and the ESL 50 questions data set. For the TOEFL data set we compare with several other methods. More specifically, we examine: the lexicon-based measures of Leacock et al. (1998) (LC), Hirst and St-Onge (1998) (HS), and Jarmasz and Szpakowicz (2003) (JS); the corpus-based measures of Landauer and Dumais (1997) (LD), Pado and Lapata (2007) (PL), Turney (2008b) (T), Terra and Clarke (2003) (TC), and Matveeva et al. (2005) (M); the hybrid measures of Resnik (1995) (R), Lin (1998) (L), Jiang and Conrath (1997) (JC), and Turney et al. (2003) (PR); and a Web-based method by Ruiz-Casado et al. (2005) (RC). We also report the results of random guessing (RG) and the performance of the average college applicant (H). Table 5 shows the results on the 80 TOEFL questions. The table reports the number of the correct and the respective percentage given by all measures. In order to test the statistical significance of the differences in the measures' performance, we conducted Fisher's Exact Test (Agresti, 1990). As in the previous tables, the symbol § indicates statistically significant difference at the 0.99 confidence level, ‡ at the 0.95 confidence level, and † at the 0.90 confidence level. The results of Table 5 show that *SR* ranks second among all reported methods, with the best method being the hybrid PR (Turney et al., 2003). With regards to its comparison with the lexicon-based methods, *SR* reports better results, statistically significant at the confidence levels indicated.

In a similar manner, we have conducted experiments in the ESL 50 questions data set, and compare our results with: the lexicon-based measures of Leacock et al. (1998) (LC), Hirst and St-Onge (1998) (HS), and Jarmasz and Szpakowicz (2003) (JS); the corpus-based measures of Turney (2001) (PMI-IR), and Terra and Clarke (2003) (TC); and the hybrid measures of Resnik (1995) (R),

| Category | Method | #Correct Answers | Percentage of Correct Answers |
|---|---|---|---|
| **Lexicon-Based** | **LC** | 17 | $0.212^\S$ |
| | **HS** | 62 | $0.775^\ddagger$ |
| | **JS** | 63 | $0.787^\dagger$ |
| **Corpus-Based** | **LD** | 52 | $0.65^\S$ |
| | **PL** | 58 | $0.725^\S$ |
| | **T** | 61 | $0.762^\S$ |
| | **TC** | 65 | $0.812^\S$ |
| | **M** | 69 | 0.862 |
| **Hybrid** | **R** | 16 | $0.2^\S$ |
| | **L** | 19 | $0.237^\S$ |
| | **JC** | 20 | $0.25^\S$ |
| | **PR** | 78 | $0.975^\ddagger$ |
| **Web-Based** | **RC** | 66 | 0.825 |
| **Other** | **RG** | 20 | $0.25^\S$ |
| | **H** | 52 | $0.65^\S$ |
| | **SR** | **70** | **0.875** |

Table 5: Number and percentage of correct answers in the TOEFL 80 questions test. Confidence levels: $^\dagger$=0.90, $^\ddagger$=0.95, $^\S$=0.99

| Category | Method | #Correct Answers | Percentage of Correct Answers |
|---|---|---|---|
| **Lexicon-Based** | **LC** | 18 | $0.36^\S$ |
| | **HS** | 31 | $0.62^\ddagger$ |
| | **JS** | 41 | 0.82 |
| **Corpus-Based** | **PMI-IR** | 37 | 0.74 |
| | **TC** | 40 | 0.8 |
| **Hybrid** | **R** | 16 | $0.32^\S$ |
| | **L** | 18 | $0.36^\S$ |
| | **JC** | 18 | $0.36^\S$ |
| **Other** | **RG** | 20 | $0.25^\S$ |
| | **SR** | **41** | **0.82** |

Table 6: Number and percentage of correct answers in the ESL 50 questions test. Confidence levels: $^\ddagger$=0.95, $^\S$=0.99

Lin (1998) (L), and Jiang and Conrath (1997) (JC). We report the results, together with random guessing, in Table 6. The results of Table 6 show that *SR* ranks first, having the same performance with JS in this data set, both outperforming all of the compared corpus-based methods. These results are very interesting, since they indicate that lexicon-based methods are very promising in the synonym identification tasks.

| Category | Method | #Correct Answers | Percentage of Correct Answers |
|---|---|---|---|
| **Lexicon-Based** | **LC** | 117 | $0.313^{\ddagger}$ |
| | **HS** | 120 | $0.321^{\dagger}$ |
| | **V** | 161 | $0.43^{\S}$ |
| **Corpus-Based** | **LRA** | 210 | $0.561^{\S}$ |
| | **PMI-IR** | 131 | $0.35^{\dagger}$ |
| **Hybrid** | **R** | 124 | $0.332^{\dagger}$ |
| | **L** | 102 | $0.273^{\S}$ |
| | **JC** | 102 | $0.273^{\S}$ |
| **Web-Based** | **B** | 150 | $0.4^{\ddagger}$ |
| **Other** | **RG** | 75 | $0.2^{\S}$ |
| | **S1** | 106 | 0.283 |
| | **S2** | 114 | 0.304 |
| | **S** | **128** | **0.342** |
| | **NB** | 142 | 0.381 |
| | **UB** | 196 | 0.524 |

Table 7: Number and percentage of correct answers in the 374 Scholastic Aptitude Test (SAT) questions. Confidence levels: $^{\dagger}$=0.90, $^{\ddagger}$=0.95, $^{\S}$=0.99

### 4.1.3 Evaluation on SAT Analogy Questions

The approach that we choose to evaluate *SR* in the analogy task is to use the typical benchmark test set employed in the related bibliography, namely the Scholastic Aptitude Test (SAT).[11] It comprises of 374 words pairs and for each target pair 5 supplementary pairs of words. The average US college applicant answered correctly only the 57 percent of the questions, and no machine-based approach has yet surpassed the performance of the average college applicant.

In Table 7, we present the number of correct answers and the respective percentage (recall) on the 374 SAT questions, of the following methods: random guessing (RG), Jiang and Conrath (1997) (JC), Lin (1998) (L), Leacock et al. (1998) (LC), Hirst and St-Onge (1998) (HS), Resnik (1995) (R), Bollegala et al. (2008) (B), Veale (2004) (V), PMI-IR (Turney, 2001) and LRA (Turney, 2006). Furthermore, we present the results of $s_1$ (Equation 11), $s_2$ (Equation 12) and $s$ (Equation 13). We also present, as before, the statistical significance of the differences in performance, conducting Fisher's exact test.

Towards the direction of combining the answers of $s_1$ and $s_2$ in a different manner than the naive average, we also report the upper bound performance of such an attempt. This is computed by simply finding the union of the correct answers that $s_1$ and $s_2$ may provide. This is reported in the table as (UB). In an effort to design a learning mechanism that would learn when to select $s_1$ or $s_2$ answers for each SAT question, with the goal to reach our upper-bound, we designed and implemented a simple representation of the SAT questions as training instances. For each SAT question, we created a training instance that has 6 features: the minimum $s_1$ value found for this question (among the five computed values for all the possible pairs), the maximum $s_1$ value, and their difference. We also added the same features regarding $s_2$. We then trained and tested a Naive Bayes classifier using ten-fold cross validation in the 374 SAT questions. The results of this experiment are shown in the table as NB (Naive Bayes). Finally, we also present the top results ever reported in the literature for the specific data set, which is the LRA method by Turney (2006). This is reported in the table as (LRA).

---

11. Many thanks to Peter Turney, for providing us with a standard set for experimentation, comprising of 374 SAT questions.

The results presented in Table 7 show that S ranks second among the compared lexicon-based measures with the first being the measure of Veale (2004) (V). The method of Bollegala et al. (2008) (B) achieves higher score than *SR*, but needs training in SAT questions. At this point we have to note that the LRA method needs almost 8 days to process the 374 SAT questions (Turney, 2006), (B) needs around 6 hours (Bollegala et al., 2008), while S needs less than 3 minutes.

Furthermore, the fact that combining $s_1$ and $s_2$ can reach $52.4\%$ shows that S can produce very promising results, if a classifier learns successfully how to combine them. The $NB$ results, which are a simple attempt to construct such a learner with few features, shows an important boost in performance of $4.1\%$. A proper feature engineering in the task, and more training SAT questions can potentially yield more promising results, as the gap between $38.1\%$ and the upper bound of $52.4\%$ is still large. In all, these results prove that our lexicon-based relatedness measure has a comparable performance to the state of the art measures for the SAT task, while it has smaller execution time than the majority of the methods which outperform it in recall.

## 4.2 Evaluation of the Omiotis Measure

In order to evaluate the performance of the *Omiotis* measure, we performed two experiments which test the ability of the measure to capture the similarity between sentences. The first experiment is based on the data set produced by Li et al. (2006). The second experiment is based on the paraphrase recognition task, using the Microsoft Research Paraphrase Corpus (Dolan et al., 2004).

### 4.2.1 EVALUATION ON SENTENCE SIMILARITY

The data set produced by Li et al. (2006) comprises 65 sentence pairs (each pair consists of two sentences that are the respective dictionary word definitions of the R&G 65 word pairs data set). The used dictionary was the Collins Cobuild dictionary (Sinclair, 2001). For each sentence pair, similarity scores have been provided by 32 human participants, ranging from $0.0$ (the sentences are unrelated in meaning), to $4.0$ (the sentences are identical in meaning).[12]

From the 65 sentence pairs, Li et al. (2006) decided to keep a subset of 30 sentence pairs, similarly to the process applied by Miller and Charles (1991), in order to retain the sentences whose human ratings create a more even distribution across the similarity range. Thus, we apply *Omiotis* in this same subset of the 65 sentence pairs, described by Li et al. (2006). In this data set, we compare *Omiotis* against the STASIS measure of semantic similarity, proposed by Li et al. (2006), an LSA-based approach described by O'Shea et al. (2008), and the STS measure proposed by Islam and Inkpen (2008). To the best of our knowledge, this data set has only been used by these three previous works. In Table 8 we present the sentence pairs used, and the respective scores by humans, STASIS, LSA, STS, and *Omiotis*.

---

12. The data set is publicly available at `http://www.docm.mmu.ac.uk/STAFF/J.Oshea/`

| Sentence Pair | Human | STASIS | LSA | STS | Omiotis |
|---|---|---|---|---|---|
| 1.cord:smile | 0.01 | 0.329 | 0.51 | 0.06 | 0.1062 |
| 5.autograph:shore | 0.005 | 0.287 | 0.53 | 0.11 | 0.1048 |
| 9.asylum:fruit | 0.005 | 0.209 | 0.505 | 0.07 | 0.1046 |
| 13.boy:rooster | 0.108 | 0.53 | 0.535 | 0.16 | 0.3028 |
| 17.coast:forest | 0.063 | 0.356 | 0.575 | 0.26 | 0.2988 |
| 21.boy:sage | 0.043 | 0.512 | 0.53 | 0.16 | 0.243 |
| 25.forest:graveyard | 0.065 | 0.546 | 0.595 | 0.33 | 0.2995 |
| 29.bird:woodland | 0.013 | 0.335 | 0.505 | 0.12 | 0.1074 |
| 33.hill:woodland | 0.145 | 0.59 | 0.81 | 0.29 | 0.4946 |
| 37.magician:oracle | 0.13 | 0.438 | 0.58 | 0.20 | 0.1085 |
| 41.oracle:sage | 0.283 | 0.428 | 0.575 | 0.09 | 0.1082 |
| 47.furnace:stove | 0.348 | 0.721 | 0.715 | 0.30 | 0.2164 |
| 48.magician:wizard | 0.355 | 0.641 | 0.615 | 0.34 | 0.5295 |
| 49.hill:mound | 0.293 | 0.739 | 0.54 | 0.15 | 0.5701 |
| 50.cord:string | 0.47 | 0.685 | 0.675 | 0.49 | 0.5502 |
| 51.glass:tumbler | 0.138 | 0.649 | 0.725 | 0.28 | 0.5206 |
| 52.grin:smile | 0.485 | 0.493 | 0.695 | 0.32 | 0.5987 |
| 53.serf:slave | 0.483 | 0.394 | 0.83 | 0.44 | 0.4965 |
| 54.journey:voyage | 0.36 | 0.517 | 0.61 | 0.41 | 0.4255 |
| 55.autograph:signature | 0.405 | 0.55 | 0.7 | 0.19 | 0.4287 |
| 56.coast:shore | 0.588 | 0.759 | 0.78 | 0.47 | 0.9308 |
| 57.forest:woodland | 0.628 | 0.7 | 0.75 | 0.26 | 0.612 |
| 58.implement:tool | 0.59 | 0.753 | 0.83 | 0.51 | 0.7392 |
| 59.cock:rooster | 0.863 | 1 | 0.985 | 0.94 | 0.9982 |
| 60.boy:lad | 0.58 | 0.663 | 0.83 | 0.60 | 0.9309 |
| 61.cushion:pillow | 0.523 | 0.662 | 0.63 | 0.29 | 0.3466 |
| 62.cemetery:graveyard | 0.773 | 0.729 | 0.74 | 0.51 | 0.7343 |
| 63.automobile:car | 0.558 | 0.639 | 0.87 | 0.52 | 0.7889 |
| 64.midday:noon | 0.955 | 0.998 | 1 | 0.93 | 0.9291 |
| 65.gem: jewel | 0.653 | 0.831 | 0.86 | 0.65 | 0.8194 |

Table 8: Human, STASIS, LSA, STS and *Omiotis* scores for the 30 sentence pairs.

In Table 9 we present the results of the comparison, comprising the reporting of the Spearman's rank order correlation coefficient $\rho$ and the Pearson's product moment correlation coefficient $r$ for STASIS, LSA, STS, and *Omiotis*. We have also included in the results, a version of *Omiotis* that does not take into account the inter-POS relations (i.e., relations that cross parts of speech). This version of *Omiotis* is indicated in the table as *SimpleOmiotis*. The objective of this experiment was to measure the contribution of the relations that cross parts of speech in the computation of text-to-text semantic relatedness values, though these types of relations have been reported in the previous bibliography as advantageous (Jarmasz, 2003; Jarmasz & Szpakowicz, 2003), but their individual contribution had never been measured.

|  | Spearman's $\rho$ | Pearson's $r$ |
|---|---|---|
| **STASIS** | 0.8126[‡] | 0.8162 |
| **LSA** | 0.8714 | 0.8384 |
| **STS** | 0.838 | 0.853 |
| **Simple Omiotis** | 0.6889[§] | 0.7277[§] |
| **Omiotis** | **0.8905** | **0.856** |
| **Average Participant** | N/A | 0.825 |
| **Worst Participant** | N/A | 0.594 |
| **Best Participant** | N/A | 0.921 |

Table 9: Spearman's and Pearson's correlations with human similarity ratings. Confidence levels: [‡]=0.95, [§]=0.99

We also show the $r$ correlation between the average participant (mean of individuals with group; $n = 32$, leave-one-out resampling and standard deviation 0.072), the worst participant (worst participant with group; $n = 32$, leave-one-out resampling) and the best participant (best participant with group; $n = 32$, leave-one-out resampling), taken from the work of O'Shea et al. (2008). In addition, we have also conducted a z-test regarding the difference between *Omiotis* correlations and the compared measures' correlations. The symbols used in the previous tables indicate the confidence level of the statistical significance. Note, also, that the reported correlations (STASIS, LSA, STS, and *Omiotis*) individually constitute statistically significant positive correlations with the human scores ($r$) and rankings ($\rho$). As the results indicate, *Omiotis* has the best correlation, according to $\rho$ and $r$ values, compared to STASIS, LSA, and STS. Furthermore, the contribution of the semantic relations that cross parts of speech is obvious, since the difference between the simple version of *Omiotis* that omits them and the defined *Omiotis* measure is large and statistically significant at the 0.99 confidence level. Overall, the results indicate that *Omiotis* can be applied successfully to the computation of similarities between small text segments, like sentences.

### 4.2.2 EVALUATION ON PARAPHRASE RECOGNITION

In order to further evaluate the performance of *Omiotis* in measuring the semantic relatedness between small text segments, we conducted additional experiments on the paraphrase recognition task using the test pairs of the Microsoft Research Paraphrase Corpus (Dolan et al., 2004). From the original data set, containing both training and test pairs, we run experiments only on the 1725 test pairs of text segments, which have been collected from news sources on the Web over a period of 18 months. For each pair, human subjects have determined whether any of the two texts in the pair consists a paraphrase of the other (direction is not an issue). The reported inter-judge agreement between annotators is 83%. The paraphrase recognition task has been widely studied in the past, since it is very important in many natural language applications, like question answering (Harabagiu

| Category | Method | Accuracy | Precision | Recall | F-Measure |
|---|---|---|---|---|---|
| **Baselines** | **Random** | 51.3 | 68.3 | 50 | 57.8 |
| | **VSM and Cosine** | 65.4 | 71.6 | 79.5 | 75.3 |
| **Corpus-based** | **PMI-IR** | 69.9 | 70.2 | 95.2 | 81 |
| | **LSA** | 68.4 | 69.7 | 95.2 | 80.5 |
| | **STS** | 72.6 | 74.7 | 89.1 | 81.3 |
| **Lexicon-based** | **JC** | 69.3 | 72.2 | 87.1 | 79 |
| | **LC** | 69.5 | 72.4 | 87 | 79 |
| | **Lesk** | 69.3 | 72.4 | 86.6 | 78.9 |
| | **L** | 69.3 | 71.6 | 88.7 | 79.2 |
| | **WP** | 69 | 70.2 | 92.1 | 80 |
| | **R** | 69 | 69 | 96.4 | 80.4 |
| | **Comb.** | 70.3 | 69.6 | 97.7 | 81.3 |
| **Machine learning-based** | **Wan et al.** | 75 | 77 | 90 | 83 |
| | **Zhang and Patrick** | 71.9 | 74.3 | 88.2 | 80.7 |
| | **Qiu et al.** | 72 | 72.5 | 93.4 | 81.6 |
| | **Finch et al.** | 74.96 | 76.58 | 89.8 | 82.66 |
| | **Omiotis** | **69.97** | **70.78** | **93.4** | **80.52** |

Table 10: *Omiotis* and competitive methods performance on the Microsoft Research Paraphrase Corpus (MSR).

& Hickl, 2006), and text summarization (Madnani, Zajic, Dorr, Fazil Ayan, & Lin, 2007). For this task we computed *Omiotis* between the sentences of every pair and marked as paraphrases only those pairs with *Omiotis* value greater than a threshold. The threshold was set to 0.2, after tuning in the training set. We used a simple approach for the tuning, namely *forward hill-climbing* and *beam search* (Guyon, Gunn, Nikravesh, & Zadeh, 2006).

We compare the performance of *Omiotis* against several other methods of various categories; more precisely, against: (a) two baseline methods, a random selection method that marks randomly each pair as being paraphrase of not (Random), and a vector-based similarity measure, using the cosine similarity measure and TF-IDF weighting for the features (VSM and Cosine) [13], (b) corpus-based methods; the PMI-IR proposed by Turney (2001), an LSA-based approach introduced by Mihalcea et al. (2006), and the STS measure proposed by Islam and Inkpen (2008), (c) lexicon-based methods; Jiang and Conrath (1997) (JC), Leacock et al. (1998) (LC), Lin (1998) (L), Resnik (1995, 1999) (R), Lesk (1986) (Lesk), Wu and Palmer (1994) (WP), and a metric that combines the measures of this category, proposed by Mihalcea et al. (2006) (Comb.), and (d) machine-learning based techniques, which also constitute the state of the art in paraphrase recognition, like the method of Wan et al. (2006), which trains a classifier with lexical and dependency similarity measures, the method of Zhang and Patrick (2005), who also build a feature vector with lexical similarities between the sentence pairs (e.g., edit distance, number of common words), the method of Qiu et al.

---

13. The features are all words of the used data set.

(2006), who use an SVM classifier (Vapnik, 1995) to decide whether or not a set of features for each sentence that has been created by parsing and semantic role labelling matches or not the respective set of the second sentence in the pair, and with what importance, and, finally, the method of Finch et al. (2005), who also train an SVM classifier based on machine translation evaluation metrics.

The results of the evaluation are shown in Table 10. The results indicate that *Omiotis* surpasses all the lexicon-based methods, and matches the combined method of Mihalcea et al. (2006). At this point we must mention that we also tuned *Omiotis* with a goal to maximize F-Measure in the test set, at the cost of dropping precision in favor of recall. This type of tuning reported an F-Measure of $81.7$, which is larger than the F-Measures of the lexicon-based, the corpus-based and two of the machine learning-based approaches. Even though the reported results used a different and simpler tuning explained previously, still the results indicate that *Omiotis* manages very well in the paraphrase recognition task and produces comparable results with the state of the art. We believe that it can be used as part of a machine learning-based method, since it is one of the best choices in lexicon-based methods for paraphrase recognition, and this also constitutes part of our plan for future work in this application.

## 5. Conclusions and Future Work

In this paper we presented a new measure of text semantic relatedness. The major strength of this measure lies in the formulation of the semantic relatedness between words. Experimental evaluation, proved that our measure approximates human understanding of semantic relatedness between words better than previous related measures. The combination of path length, nodes' depth and edges' type in a single formula allowed us to apply our semantic relatedness measure to different text-based tasks with very good performance. More specifically, the *SR* measure outperformed overall in the used data sets all state of the art measures in word-to-word tasks and the *Omiotis* measure performed very well in the sentence similarity and the paraphrase recognition tasks. Although, the results in the word analogy task are satisfactory, since no special tuning has been performed, we are confident that there is still place for improvement. The extensive evaluation of *SR* and *Omiotis* in several applications shows the capabilities of our measures and proves that both can be applied to several text related tasks. It is on our next plans to apply our relatedness measures to more applications, such as text classification and clustering, keyword and sentence extraction, and query expansion, and compare with state of the art techniques in each field. Finally, we are improving our supporting infrastructure in order to facilitate large scale tasks such as document clustering and text retrieval.

## Acknowledgments

Part of this work was done while George Tsatsaronis was at the Department of Informatics of Athens University of Economics and Business. We would like to thank Kjetil Nørvåg for his constructive comments, and Ion Androutsopoulos for his feedback on the early stage of this work. We would also like to thank the anonymous reviewers for their detailed feedback.